\pdfoutput=1

\documentclass[11pt,table]{article}
\usepackage{EMNLP2022}
\usepackage{times}
\usepackage{latexsym}

\usepackage[T1]{fontenc}

\usepackage[utf8]{inputenc}

\usepackage{microtype}
\usepackage{xcolor}
\usepackage{inconsolata}
\usepackage{amsmath}
\usepackage{amssymb}
\usepackage{mathtools}
\usepackage{amsthm}
\usepackage{tabularx}
\usepackage[ruled,vlined]{algorithm2e}

\newlength\mylength
\newlength\mylengthl
\newlength\mylengths
\title{SQuAT: Sharpness- and Quantization-Aware Training for BERT}

\author{Zheng Wang$^*$, Juncheng B Li\thanks{\ \ Co-first author}\ ,       Shuhui Qu, Florian Metze, Emma Strubell\\
\{zhengwan, junchenl, fmetze, strubell\}@cs.cmu.edu}

\begin{document}
\maketitle
\begin{abstract}
Quantization is an effective technique to reduce memory footprint, inference latency, and power consumption of deep learning models. 
However, existing quantization methods suffer from accuracy degradation compared to full-precision (FP) models due to the errors introduced by coarse gradient estimation through non-differentiable quantization layers. 
The existence of sharp local minima in the loss landscapes of overparameterized models (e.g., Transformers) tends to aggravate such performance penalty in low-bit (2, 4 bits) settings. 
In this work, we propose sharpness- and quantization-aware training (SQuAT), which would encourage the model to converge to flatter minima while performing quantization-aware training. 
Our proposed method alternates training between sharpness objective and step-size objective, which could potentially let the model learn the most suitable parameter update magnitude to reach convergence near-flat minima.
Extensive experiments show that our method can consistently outperform state-of-the-art quantized BERT models under 2, 3, and 4-bit settings on GLUE benchmarks by 1\%, and can sometimes even outperform full precision (32-bit) models.
Our experiments on empirical measurement of sharpness also suggest that our method would lead to flatter minima compared to other quantization methods. 
\end{abstract}
\begin{figure}[t]
    \centering
    \includegraphics[width=\linewidth]{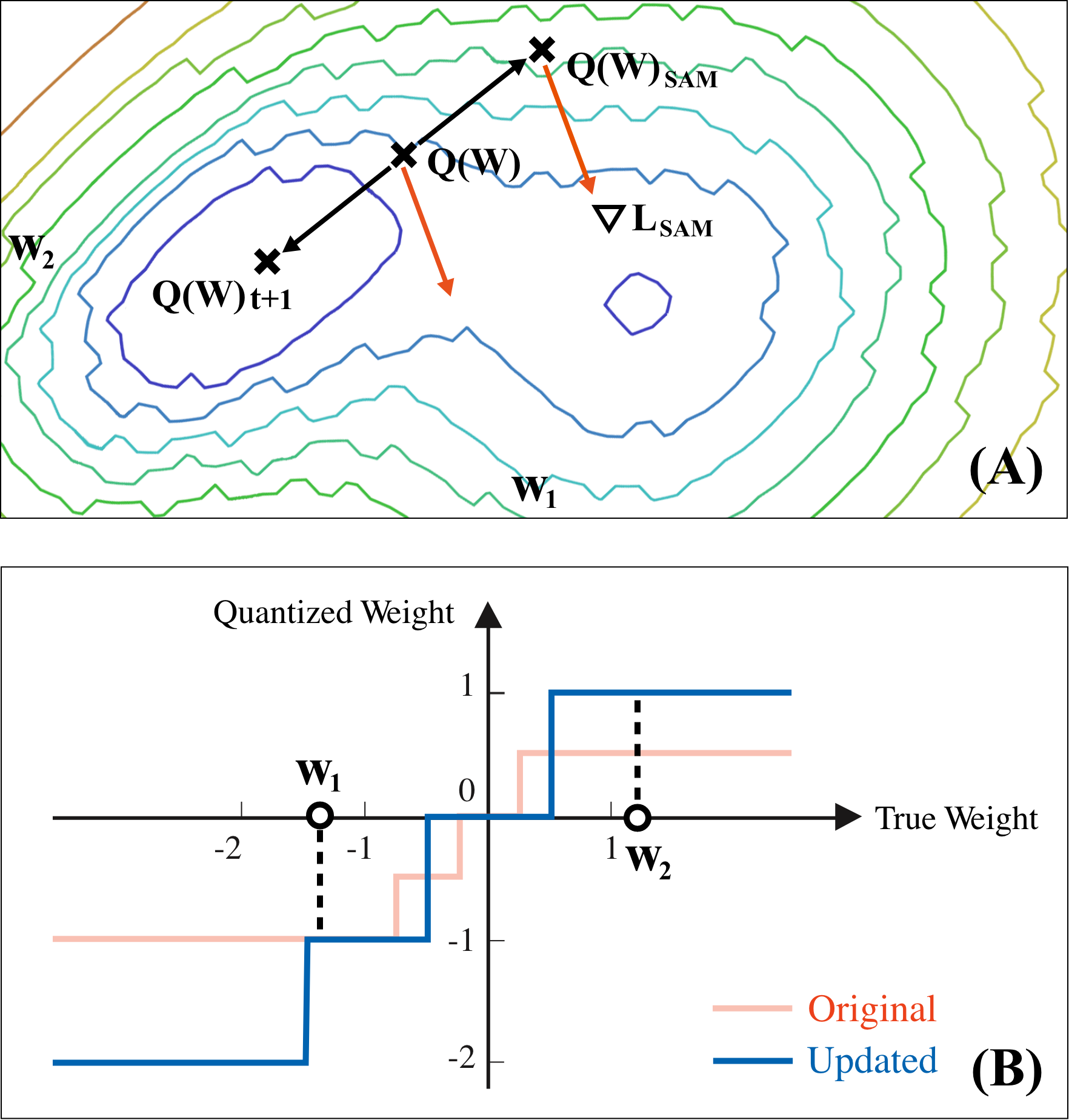}
    \caption{Illustration of our SQuAT alternate training in a 2D toy example, where $W_1, W_2$ indicate the 2 dimensions. (A) when step-size is fixed, update weights in the Sharpness-aware minimizing (SAM) direction in the quantized loss landscape. (B) Alternating step: fix weight, update the step-size. }
    \label{fig:overall}
\end{figure}
\section{Introduction}
As state-of-the-art deep learning models for NLP and speech (e.g. Transformers, BERT) grow increasingly large and computationally burdensome \cite{devlin2018bert, karita2019comparative}, there is increasing antithetical demand, motivated by latency and energy consumption concerns, to develop computationally efficient models.
Model quantization has emerged as a promising approach to enable the training and deployment of deep learning models with high compression rates, resulting in a significant reduction in memory bandwidth and less carbon footprint. 

There are two primary strategies for quantization: \emph{Post-training} quantization (PTQ) quantizes the parameters of a model trained in full precision post hoc and tends to suffer a heavy penalty on the accuracy, as its inference graph differs substantially from the training~\cite{jacob2018quantization}. \emph{Quantization-aware training} (QAT)~\cite{pulkitQAT} combats this discrepancy by simulating quantization during training: All weights are quantized to low bit-width during both forward and backward passes of training, so that the model parameters will work well when inference is performed in low bit-width. 
Note that in the backward pass, QAT leverages straight through estimator~(STE; \cite{bengio2013estimating}) to estimate the gradient of these non-differentiable quantizer layers.

To date, QAT of BERT models still introduces a non-negligible performance loss compared to the full-precision models~\cite{zafrir2019q8bert,bai2020binarybert}.
This drop in accuracy is caused by poor estimation of gradients due to the STE-step through nondifferentiable layers. 
To alleviate errors caused by such coarse estimation, LSQ~\citep{esser2019learned} proposed to have the quantization bin width, or \emph{step size}, update proportional to the weights update during training. 
Such finer grain estimation achieves impressive performance when quantizing ResNet models~\cite{esser2019learned}, but suffers heavier accuracy loss when quantizing larger Transformer models. 
This performance gap between quantized ResNets and Transformers could be explained by the different geometry of their loss surface, where Transformers were shown to have a much sharper loss landscape than ResNets~\cite{chen2021vision}. 

In the meantime, the line of research that focuses on the flatness of minima has garnered increasing interest.
Strong empirical evidence shows that SAM~\cite{foret2020sharpness} optimization can improve model generalization by simultaneously minimizing loss value and loss sharpness. Models trained with this objective achieved better generalization across many tasks in the NLP and vision domain~\cite{chen2021vision, mehta2021empirical}.

In this work, we incorporate the intuition of SAM and combine it with step-size quantization. 
However, the combination of the two strategies is nontrivial: naively merging the sharpness term and step-size optimization and jointly optimizing the loss function results in unstable performance, in some cases underperforming LSQ. 
The reason is explained in \S~\ref{sec:alternate}. 
To overcome such instability, we introduce an alternate training schedule SQuAT, which optimizes for the step size in one pass and switches gear to optimize according to the SAM objective in the next pass, and keeps alternating until the convergence of the model.

In our experiments, we fine-tune a pre-trained uncased BERT baseline model on each GLUE benchmark~\cite{wang2018glue} task using our SQuAT quantization.
The result shows that our method achieves about 1\% performance gain over SOTA quantized models consistently across 8 different tasks. Our contributions are listed as follows:
\begin{enumerate}
    \item We successfully leveraged an alternate training schedule to let LSQ work stably together with SAM optimization, in contrast to a naive joint training method that leads to unstable performance (sometimes worse than LSQ alone). Both are pioneering efforts.
    
    \item Our experiments suggest that our SQuAT quantization achieves a sizeable performance gain over other state-of-the-art quantization methods at low bitwidth.
    
    \item Measurement of sharpness verifies that the models trained with our SQuAT indeed converge into flatter minima compared to LSQ. 
\end{enumerate}
\section{Background \& Related works}
\subsection{Learnable parameters Quantization}
Although the community sees some renewed interest in post-training quantization (PTQ), it still lags behind quantization-aware training (QAT) methods in accuracy, as is concluded by~\citet{gholami2021survey}.
Hence, in this work, we focus on QAT, which was first introduced by~\citet{jacob2018quantization}.
Several works proposed learning-based approaches to improve QAT around the same time: \citet{jain2019trained} proposed to learn the quantizers’s dynamic range while training; \citet{uhlich2019mixed} advocated for learning the optimal bit-width. 
Among them, LSQ~\cite{esser2019learned} was the simplest and best performing approach, which proposes learning the optimal quantizer step size (bin width).
In their work, a trainable scale parameter $s$ is proposed for both weights and activations. This scheme defines the \emph{elementwise} quantization function as:
\begin{equation}
    \mathbf{Q}(w) = \lfloor clip(w/s, Q_N, Q_P) \rceil \cdot s
    \label{eqn:lsq}
\end{equation}
where $\lfloor \cdot \rceil$ indicates rounding and the $clip(\cdot)$
function clamps all values between $Q_N = -2^{n-1}$ and $Q_P = -2^{n-1}-1$ in a $n$ bit setting. The parameter $w$ is the weight value to be quantized and $s$ is the learned scaling \emph{scalar} (one per module). Following the rules of STE, this quantizer naturally results in a step-size gradient of:
\begin{small}
\begin{equation}
    \frac{\partial \mathbf{Q}(w)}{\partial s}=\begin{cases}
        -w/s + \lfloor w/s \rceil & \text{if $-Q_N < w < Q_P$}\\
        -Q_N & \text{if $w \le -Q_N$}\\
        Q_P & \text{if $w \ge Q_P$}
    \end{cases}
    \label{s-grad}
\end{equation}
\end{small}

\begin{table*}[ht!]
\scriptsize
\begin{tabular}{p{\mylength}clp{\mylengths}p{\mylengthl}p{\mylengthl}p{\mylengthl}p{\mylengthl}p{\mylengths}p{\mylengths}}
\textbf{Task}    & \textbf{W/A Bits}     & \textbf{COLA}                      & \textbf{SST-2}                     & \textbf{MRPC}              & \textbf{STS-B}             & \textbf{QQP}               & \textbf{mNLI}              & \textbf{qNLI}                      & \textbf{RTE}                       \\ 
\textbf{Metrics}& & Matthews Corr. & Acc.  & Acc. & Peason Corr. &Acc. &Matched Acc. & Acc. & Acc.
\\\hline
FP32 & 32/32 & \multicolumn{1}{l}{56.5} & \multicolumn{1}{l}{93.1} & 82.84       & 88.6         & 90.8                           & 84.34                          & \multicolumn{1}{l}{91.4} & \multicolumn{1}{l}{67.2} \\
\hline
Q8BERT & 8/8 & $\text{58.5}_{\pm1.3}$ & $\text92.2_{\pm0.3}$& - & $\text89.0_{\pm0.2}$ & $\text88.0_{\pm0.4}$ & - &$\text90.6_{\pm0.3}$ & $\text68.8_{\pm3.5}$\\  
\hline
GOBO & 2/32 & - & - & - & 82.7  & - & 71.0 & - & -\\ 
Q-BERT & 2/8 & - & 84.6 &-&-&-&76.6&-&-\\  
LSQ* &2/8  & $\text51.3_{\pm0.7}$                                         & $\text92.2_{\pm0.1}$                                         & $\text83.5_{\pm0.7}$ & $\text87.2_{\pm0.1}$ & $\text91.1_{\pm0.1}$                   & $\text83.6_{\pm0.1}$                    & $\text91.1_{\pm0.1}$                                         & $\text66.8_{\pm0.9}$                                         \\
\textbf{SQuAT}* & 2/8  & $\textbf{53.3}_{\pm0.2}$ & $\textbf{92.7}_{\pm0.1}$ & $\textbf{84.0}_{\pm0.7}$ & $\textbf{88.0}_{\pm0.1}$ & $\textbf{91.1}_{\pm0.2}$ & $\textbf{84.0}_{\pm0.1}$& $\textbf{91.3}_{\pm0.1}$ & $\textbf{67.4}_{\pm0.4}$\\
\hline{}
GOBO & 3/32 & - & - & - & 88.3 & - & 83.7 & - & -\\ 
Q-BERT & 3/8 &- & 92.5 &-&-&-&83.4&-&-\\  
LSQ* & 3/8   & $\text58.8_{\pm0.4}$                                         & $\text92.6_{\pm0.2}$                                        & $\text84.2_{\pm0.5}$ & $\text88.4_{\pm0.1}$ & $91.2_{\pm0.2}$                   & $\text84.2_{\pm0.2}$                  & $\text91.8_{\pm0.1}$                                         & $\text68.6_{\pm0.3}$                                         \\
\textbf{SQuAT}* & 3/8  & $\textbf{59.2}_{\pm0.6}$ & $\textbf{93.0}_{\pm0.2}$ & $\textbf{86.2}_{\pm0.7}$ & $\textbf{89.1}_{\pm0.1}$ & $\textbf{91.5}_{\pm0.1}$ & $\textbf{84.6}_{\pm0.1}$& $\textbf{92.1}_{\pm0.1}$ & $\textbf{70.6}_{\pm0.6}$\\
\hline
Q-BERT & 4/8 &- & 92.7 &-&-&-&83.9&-&-\\
LSQ* & 4/8  & $\text{58.7}_{\pm0.5}$ & $\text{93.2}_{\pm0.1}$ & $\text{83.4}_{\pm0.6}$ & $\text{89.1}_{\pm0.1}$ & $\text{91.5}_{\pm0.1}$ & $\text{84.6}_{\pm0.1}$& $\text{91.6}_{\pm0.1}$ & $\text{68.6}_{\pm0.4}$\\
\textbf{SQuAT}* & 4/8  & $\textbf{59.1}_{\pm0.4}$ & $\textbf{93.6}_{\pm0.2}$ & $\textbf{85.1}_{\pm0.5}$ & $\textbf{89.3}_{\pm0.1}$ & $\textbf{91.5}_{\pm0.1}$ & $\textbf{85.1}_{\pm0.1}$& $\textbf{91.9}_{\pm0.2}$ & $\textbf{69.4}_{\pm0.7}$
\end{tabular}
\caption{Performance comparisons of different quantization methods on GLUE benchmark. The score is evaluated on the development set of the task using the specified metric. We compare our SQuAT against GOBO~\cite{zadeh2020gobo} (PTQ baseline),  Q-BERT~\cite{shen2020q}, and LSQ~\cite{esser2019learned}. We also list full-precision model (FP32), and the 8-bits Q8Bert~\cite{zafrir2019q8bert} model as references. We report the mean and standard deviation of the performance calculated over 3 random seeds. "-" denotes results were not reported in the original paper. Note here ~\cite{bai2020binarybert, zhang2020ternarybert, kim2021bert} all initialized with different BERT models than our BERT$_{base}$ model, and thus are not listed in comparison here. For tasks with multiple metrics, we report the main metric here. The alternative metrics are shown in Table \ref{tab:auxi} of the appendix. * indicates our own implementation}
\label{tab:result}
\end{table*}

\subsection{Quantization of BERT}
Since training BERT usually involves adaptive optimizers like Adam rather than SGD, and BERT landscape is shown to be sharper than ResNet~\cite{chen2021vision}, many QAT methods that worked with ResNet on vision tasks would not perform equally well with the heavier parameterized BERT on the NLP benchmark~\cite{gholami2021survey}. 
GOBO~\cite{zadeh2020gobo} is a recent benchmark for PTQ in BERT. 
For QAT baselines, there are \citet{zafrir2019q8bert} and Q-BERT~\cite{shen2020q}.
BinaryBERT~\cite{bai2020binarybert} and Tenary BERT~\cite{zhang2020ternarybert} attempted the challenging ``binarization" task of BERT models. The most recent effort is to perform integer quantization of the RoBERTa model~\cite{kim2021bert}.

\subsection{Flat Minima}
A flat minimum in the loss landscape is a local optima where the loss remains low in a nearby region. 
We follow the $\epsilon$-sharpness definition of \cite{keskar2016large} which defines sharpness as maximum loss within a neighborhood bounded by $\epsilon$. In math expression, let $\boldsymbol{w}$ denotes the collection of all model weights, $\max_{\|\epsilon \|_2 < \rho} \mathcal{L}(\boldsymbol{w} + \epsilon) - \mathcal{L}(\boldsymbol{w})$ is small under a given radius, $\rho$. 

\looseness=-1 In order to achieve a flatter optima, SAM \cite{foret2020sharpness} introduces a minimax objective: $\min_{\boldsymbol{w}} \max_{\|\epsilon \|_2 < \rho} \mathcal{L}(\boldsymbol{w}+\epsilon)$
to push models into flat minima and proposes the following gradient update under $\ell_2$ norm (Note that to reduce computational cost, $\epsilon(w)$ in Eq.~\ref{sam_update} is regarded as constant and no gradient flows to it):
\begin{equation}
\begin{split}
    \epsilon(\boldsymbol{w}) \approx \rho \nabla_{\boldsymbol{w}}\mathcal{L}(\boldsymbol{w})/\|\nabla_{\boldsymbol{w}}\mathcal{L}(\boldsymbol{w})\|_2\\
    \boldsymbol{w} \leftarrow \boldsymbol{w} - \eta\cdot \nabla_{\boldsymbol{w}} \mathcal{L}(\boldsymbol{w}+\epsilon(\boldsymbol{w}))
\end{split}
\label{sam_update}
\end{equation}
 \citet{nahshan2020loss} pioneered the effort to let quantization be aware of loss landscape.
A more recent unpublished work~\cite{liu2022sharpness}(concurrent to our SQuAT), which also leverages SAM to perform quantization adopts manually defined step-size and only ran experiments on vision datasets.

\begin{figure*}[t]
    \centering
    \includegraphics[width=0.82\linewidth]{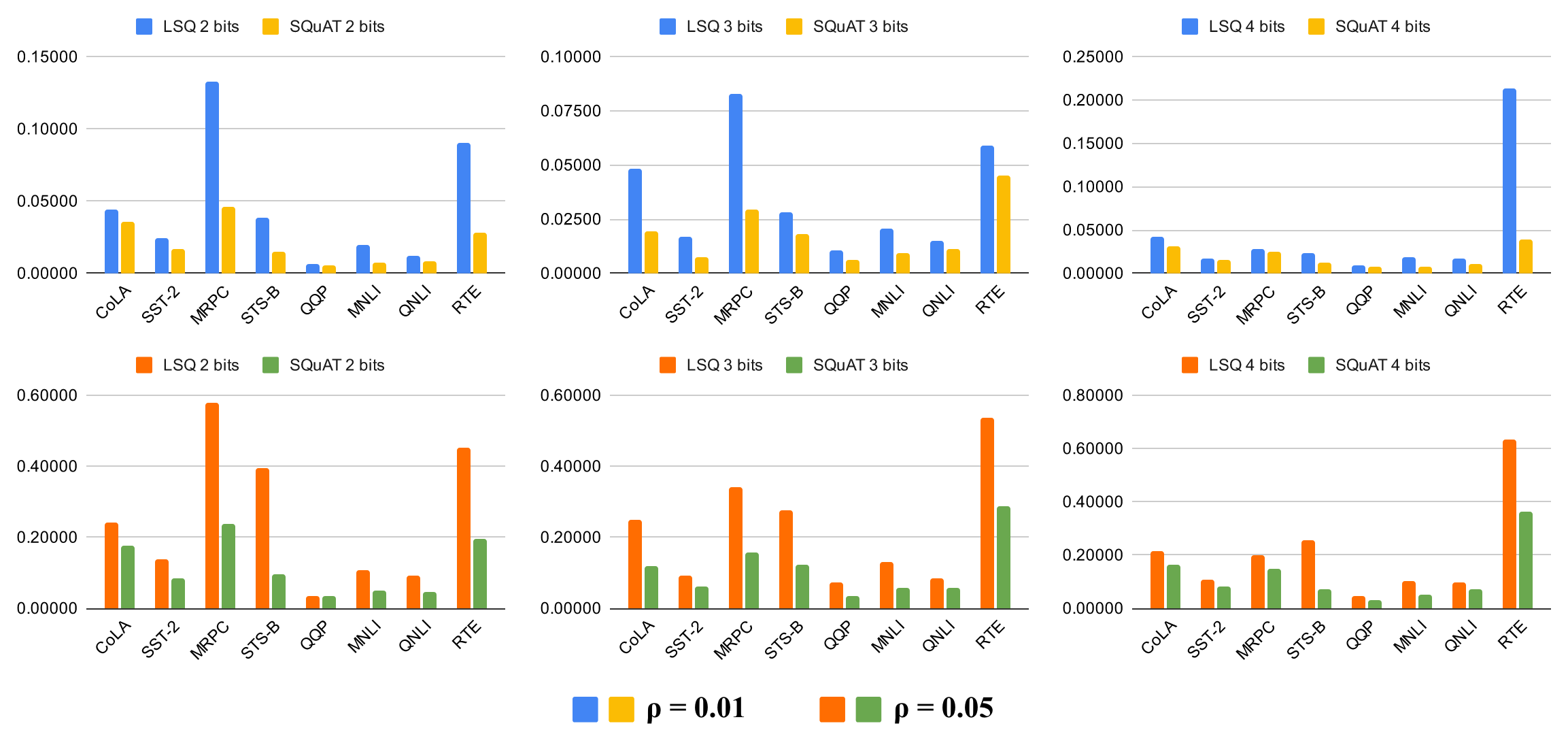}
    \caption{Sharpness of SQuAT VS. LSQ for all GLUE Tasks. The lower sharpness means a flatter local minima.}
    \label{fig:sharpness}
    \vspace{-0.6cm}
\end{figure*}
\section{Methodology}
\label{sec:alternate}
The most intuitive way to apply SAM to the quantized model is to optimize for the following objective, where $\boldsymbol{w}$ is the collection of model weights and $\boldsymbol{s}$ the collection of step-size parameters:
\begin{equation}
    \min_{\boldsymbol{w},\boldsymbol{s}} \max_{\|\epsilon \|_2 < \rho} \mathcal{L}(\mathbf{Q}(\boldsymbol{w},\boldsymbol{s})+\epsilon)
    \label{q_minimax}
\end{equation}
We add the SAM perturbation directly to the quantized weight because the element-wise gradient from $\mathbf{Q}(w,s)$ to $w$ and $s$ can only be estimated through STE, therefore the SAM perturbation, $\epsilon$, cannot be accurately evaluated. 
A natural approach to optimize Eq.~\ref{q_minimax} is to jointly update $\boldsymbol{s}$ and $\boldsymbol{w}$, which results in the following updates: 
\begin{small}
\begin{align}
    \epsilon \approx \rho \nabla_{\mathbf{Q}}\mathcal{L}(\mathbf{Q}(\boldsymbol{w},\boldsymbol{s}))/\|\nabla_{\mathbf{Q}}\mathcal{L}(\mathbf{Q}(\boldsymbol{w},\boldsymbol{s}))\|_2\\
    \boldsymbol{s} \leftarrow \boldsymbol{s} - \eta \nabla_{\mathbf{Q}} \mathcal{L}(\mathbf{Q}(\boldsymbol{w},\boldsymbol{s})+\epsilon) \nabla_{\boldsymbol{s}}\mathbf{Q}(\boldsymbol{w},\boldsymbol{s}) \label{s_up}\\
    \boldsymbol{w} \leftarrow \boldsymbol{w} - \eta \nabla_{\mathbf{Q}} \mathcal{L}(\mathbf{Q}(\boldsymbol{w},\boldsymbol{s})+\epsilon) \label{q_up}
\end{align}
\end{small}

\begin{figure}[t]
    \centering
    \includegraphics[width=0.8\linewidth]{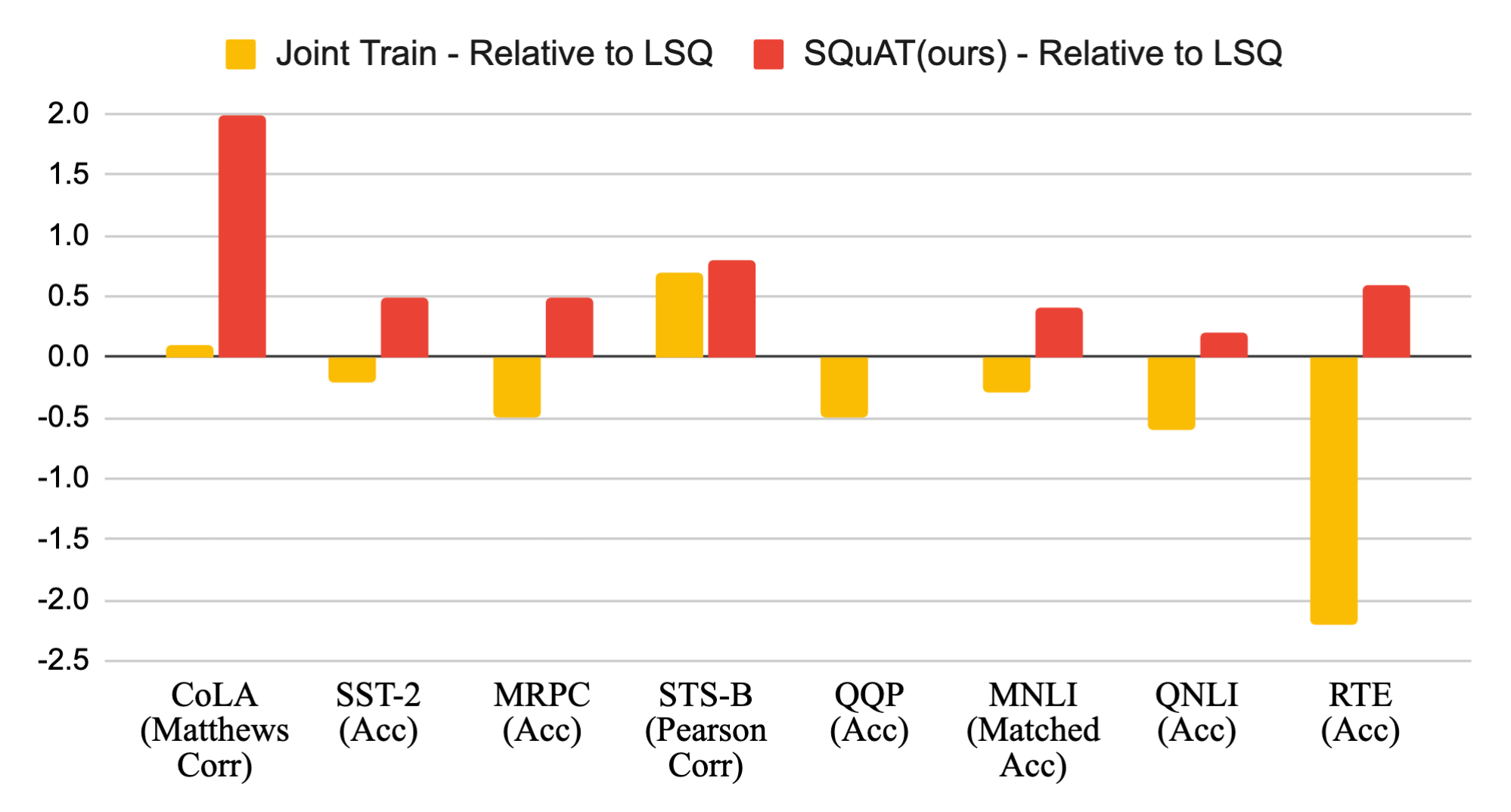}
    \caption{Joint Training VS. Alternate Training performance difference relative to LSQ, which is the 0 line}
    \label{fig:joint}
    \vspace{-0.6cm}
\end{figure}

Notice that we are omitting the term $\nabla_{\boldsymbol{w}}\mathbf{Q}(\boldsymbol{w},\boldsymbol{s})$ when updating $\boldsymbol{w}$ since for each element being updated, the STE gradient $\partial \mathbf{Q}(w,s)/\partial w=1$ .
If we take $\mathbf{Q}(\boldsymbol{w}, \boldsymbol{s})\rightarrow \boldsymbol{w}$, Eq.~\ref{q_up} breaks down to the update in Eq. \ref{sam_update}. 
However, this is not the case for step-size parameter $\boldsymbol{s}$ (Eq. \ref{s_up}). The joint update of each step-size parameter has elementwise gradient $\partial \mathbf{Q}(w,s)/ \partial s$ which must be evaluated at run-time.
Thus the joint update of step-size cannot approximate the effect of Eq. \ref{sam_update} similar to the weight update.
This makes the update of step-size asymmetric w.r.t the update of model weight $\boldsymbol{w}$ and intuitively, may not be able to catch up with the weight updates in the sharpness-aware direction.
To fix this, we make the weight updates to ``wait'' for the step-size to adapt to the proper magnitude.
Specifically, we inherit the sharpness-aware update of $\boldsymbol{w}$, while during the step-size updating phase, we fix the model weights $\boldsymbol{w}$, and only update $\boldsymbol{s}$, as summarized in Algorithm \ref{algo:main}. Because the weights are already sharpness-aware terms, we also simplified updates of the highly shared step-sizes term to regular SGD.
Comparatively, our algorithm exhibits better stability, which is corroborated by our strong empirical results in Table \ref{tab:result}.
Empirically, we also observe joint update resulting in lower performance compared to alternate training as shown in Figure~\ref{fig:joint}, with joint training underperforming LSQ baseline in majority of the GLUE~\cite{wang2018glue} tasks.

\begin{algorithm}[h]
\nl \While{not converging}{
$\rightarrow$ sample Batch B;\\
$\rightarrow$ compute gradient $\nabla_{\mathbf{Q}}\mathcal{L}(\mathbf{Q}(\boldsymbol{w}_t,\boldsymbol{s}_t))$\\
$\rightarrow$ compute $\epsilon(\mathbf{Q}(\boldsymbol{w}_t,\boldsymbol{s}_t))$\\
$\rightarrow$ update $\boldsymbol{w}$ with STE gradient $\boldsymbol{w}_{t+1} \leftarrow \boldsymbol{w}_t - \eta \cdot  \nabla_{\mathbf{Q}} \mathcal{L}(\mathbf{Q}(\boldsymbol{w}_t, \boldsymbol{s}_t)+\epsilon)$\\
$\rightarrow$ update $\boldsymbol{s}$ w.r.t weights $\boldsymbol{s}_{t+1} \leftarrow \boldsymbol{s}_{t}+\eta \cdot \nabla_{\boldsymbol{s}}\mathcal{L}(\mathbf{Q}(\boldsymbol{w}_{t+1},\boldsymbol{s}_t))$\\
}
\caption{\label{algo:main}Alternate Training}
\end{algorithm}

\section{Experiment \& Discussion}

We apply SQuAT to quantize the pre-trained uncased BERT baseline model and evaluate the performance of our proposed quantization on the GLUE benchmark~\cite{wang2018glue}, which consists of a collection of NLP tasks. 
For all these tasks, we run the experiments with three random seeds and report the mean and standard deviation of the result\footnote{More details of the setup is included in the appendix} 
As shown in Table\ref{tab:result}, our SQuAT outperforms all existing quantization methods in all GLUE tasks under the 2,3 and 4 bits scheme. 
SQuAT significantly outperforms GOBO\cite{zadeh2020gobo} by at least 5\% in 2 bits and 1\% in 3 bits in GLUE tasks, which shows the necessity of quantization-aware training (QAT) over PQT.
We compare SQuAT with other QAT methods, including Q-BERT\cite{shen2020q} and the current SOTA LSQ\cite{esser2019learned}.
When quantizing to 2-bit, 3-bit, and 4-bit, BERT model quantized with SQuAT outperforms LSQ quantized BERT by 2\% on average for all 8 GLUE tasks.
Remarkably, our 3-bit and 4-bit performance exceed the full-precision score by 1\% at several GLUE tasks. 

To show that the SQuAT quantization method reduces the sharpness of the minima, we measure the sharpness score of the local minimum in the loss landscape following~\cite{foret2020sharpness,mehta2021empirical}. The result is shown in Figure.~\ref{fig:sharpness} \footnote{Refer to Appendix for how we computed the score, and more details in Table.~\ref{tab:rho001} \& \ref{tab:rho005}}. Compared to LSQ, we observe that the models trained with our SQuAT quantization converge to much flatter minima across all GLUE tasks. 


\section{Limitation}
SQuAT method optimizes for the sharpness-aware perturbation and step size alternatively, and thus would incur more computational overheads and potentially longer train time compared to the classical QAT methods, but there is no additional cost in inference time. 
Inevitably, as a method of QAT, we inherit the limitations of QAT, which is ``fake" quantization, meaning all computations are still done with full-precision floating point numbers.
This approach is more expensive to carry out in practice than post training quantization, but generates better results. 
\clearpage
\bibliography{anthology,custom}
\bibliographystyle{acl_natbib}

\clearpage
\appendix

\section{Unform VS. non-Uniform Quantization}
Quantization approaches can be subdivided into uniform and non-uniform quantization. Non-uniform quantization tends to achieve better accuracy than uniform quantization~\cite{yamamoto2021learnable}, but requires nonstandard hardware support to store codebooks or quantization intervals, hence not practical with existing hardware~\cite{gholami2021survey}. 
In the scope of this paper, we refer QAT as \emph{uniform} quantization via quantization-aware training.

\section{Experiment Setup}
We initialized the model with uncased BERT$_{base}$ model~\cite{wolf2019huggingface} from the HuggingFace library with pre-trained weights, and fine-tuned on each GLUE benchmark task. 
We follow the same setup as \cite{chen2020lottery} and report validation set accuracy for QQP, QNLI, MRPC, RTE, SST-2, matched accuracy for MNLI, Matthew’s correlation for CoLA, and Pearson correlation for STS-B, the alternative metrics are included below in Table\ref{tab:auxi}.
Each task was trained on its train set for 5 epochs to obtain the starting checkpoint for the all of the quantization model. 
To train the quantization model, we use Adam optimizer with initial learning rate set at 1e-5 and use cosine annealing LR schedule to adjust the learning rate during the training process. 
To perform the SQuAT and LSQ fine-tuning, we run each model for 32 epochs for each tasks.
The hyperparameter $\rho$ we used for training SQuAT is 0.1 for 2-bits and 3-bits models, and 0.15 for 4-bits models, which are determined by a grid search from 0.0 to 0.25 at 0.05 increment on MNLI task.

\section{Measuring sharpness}
Following~\citet{mehta2021empirical}, Algorithm \ref{algo:sharpness} shows how we compute the sharpness score for our quantized model checkpoints. $W_t$ is the weight at $T$ step, $\eta$ is learning rate, $\rho$ is a small radius. $\nabla_W$ is the gradient, $\mathcal{L}$ is the task loss function (in our case, differs at different GLUE~\cite{wang2018glue} task).
\begin{algorithm}[h]
\nl buffer initial weight as $w_0$\\
\nl \While{not converging}{
\eIf{$\|W_{t+1}-W_0\|_2 \le \rho$}{
    $W_{t+1} = W_{t} + \eta \cdot \nabla_W \mathcal{L}$
  }{
    $W_{t+1} = W_{t} + \eta \cdot \nabla_W \mathcal{L}$\\
    $W_{t+1} = \rho\cdot \frac{W_{t+1}-W_0}{\|W_{t+1}-W_0\|_2} + W_0$
  }
$t = t+1$
}
\nl Return $\mathcal{L}(W_{T+1}) - \mathcal{L}(W_{0})$;
\caption{\label{algo:sharpness} Sharpness Measurement}
\end{algorithm}

\begin{table*}[t]
\small
\centering
\begin{tabular}{p{\mylength}cp{\mylengths}cccc}
\textbf{Task}    & \textbf{Bits}                        & \textbf{MRPC}              & \textbf{STS-B}             & \textbf{QQP}               & \textbf{mNLI}          \\ 
\textbf{Metrics}& & F1 & Spearman corr. & F1 & Mismatched Acc. \\
\hline
FP32 & 32 & 88.01       & 88.2         & 87.7                           & 84.53 \\
\hline
Q8BERT & 8 & $\text89.6_{\pm0.2}$ & - & - & - \\  
\hline
Q-BERT & 2 &-&-&-&77.0\\
LSQ &2 & $\text88.6_{\pm0.4}$ & $\text86.7_{\pm0.1}$ & $\text88.0_{\pm0.1}$                   & $\text83.6_{\pm0.1}$\\
\textbf{SQuAT} & 2 & $\textbf{88.9}_{\pm0.5}$ & $\textbf{87.4}_{\pm0.1}$ & $\textbf{88.0}_{\pm0.1}$ & $\textbf{84.1}_{\pm0.2}$\\
\hline
Q-BERT & 3 &-&-&-&83.8\\
LSQ & 3   & $\text88.7_{\pm0.3}$ & $\text87.9_{\pm0.1}$ & $88.4_{\pm0.1}$                   & $\text84.0_{\pm0.2}$ \\
\textbf{SQuAT} & 3  & $\textbf{90.6}_{\pm0.5}$ & $\textbf{88.6}_{\pm0.1}$ & $\textbf{88.6}_{\pm0.1}$ & $\textbf{84.7}_{\pm0.1}$\\
\hline
Q-BERT & 4 &-&-&-&84.2\\
LSQ & 4  & $\text{88.4}_{\pm0.4}$ & $\text{88.7}_{\pm0.1}$ & $\text{88.6}_{\pm0.1}$ & $\text{84.7}_{\pm0.2}$\\
\textbf{SQuAT} & 4  & $\textbf{89.5}_{\pm0.5}$ & $\textbf{88.8}_{\pm0.1}$ & $\textbf{88.6}_{\pm0.1}$ & $\textbf{84.8}_{\pm0.1}$

\end{tabular}
\caption{GLUE benchmark results on auxiliary metric. We report mean and standard deviation calculated over 3 random seeds.}
\label{tab:auxi}
\end{table*}

\begin{table*}[t]
\centering
\small
\begin{tabular}{llllllllll}
\textbf{Task}  & \textbf{Bits}     & \textbf{COLA}    & \textbf{SST-2}    & \textbf{MRPC}    & \textbf{STS-B}    & \textbf{QQP}     & \textbf{mNLI }   & \textbf{qNLI }   & \textbf{RTE }    \\ \hline
LSQ &2 & 0.04393 & 0.02461 & 0.13278 & 0.03866 & 0.00659 & 0.01969 & 0.01189 & 0.09071 \\
SQuAT &2 & \textbf{0.03564} & \textbf{0.01656} & \textbf{0.04578} & \textbf{0.01524} & \textbf{0.00533} & \textbf{0.00783} & \textbf{0.00833} & \textbf{0.02861} \\
\hline
LSQ &3 & 0.04835 & 0.01660 & 0.08314 & 0.02848 & 0.01046 & 0.02036 & 0.01520 & 0.05867 \\
SQuAT &3 & \textbf{0.01968} & \textbf{0.00773} & \textbf{0.02964} & \textbf{0.01789} & \textbf{0.00626} & \textbf{0.00905} & \textbf{0.01111 }& \textbf{0.04527} \\
\hline
LSQ &4 & 0.04174 & 0.01783 & 0.02867 & 0.02268 & 0.00850 & 0.01851 & 0.01771 & 0.21308 \\
SQuAT &4 & \textbf{0.03092 }& \textbf{0.01594} & \textbf{0.02517 }& \textbf{0.01232 }& \textbf{0.00715 }&\textbf{ 0.00805} & \textbf{0.01154 }& \textbf{0.03905}
\end{tabular}
\caption{The Sharpness comparison between LSQ and SQuAT on GLUE benchmark. Here, $\rho=0.01$. The local minimum of SQuAT is clearly flatter than LSQ by a large margin across all the GLUE tasks.}
\label{tab:rho001}
\end{table*}

\begin{table*}[t]
\centering
\small
\begin{tabular}{llllllllll}
\textbf{Task}  & \textbf{Bits}     & \textbf{COLA}    & \textbf{SST-2}    & \textbf{MRPC}    & \textbf{STS-B}    & \textbf{QQP}     & \textbf{mNLI }   & \textbf{qNLI }   & \textbf{RTE }    \\ \hline
LSQ &2  & 0.23983 & 0.13650 & 0.57738 & 0.39474 & 0.03593 & 0.10609 & 0.09377 & 0.45210 \\
SQuAT &2  & \textbf{0.17667} & \textbf{0.08475 }& \textbf{0.23903} & \textbf{0.09545} & \textbf{0.03421} & \textbf{0.04913} & \textbf{0.04665} & \textbf{0.19441} \\
\hline
LSQ &3  & 0.25096 & 0.09075 & 0.34070 & 0.27663 & 0.07237 & 0.13044 & 0.08277 & 0.53654 \\
SQuAT &3  & \textbf{0.11754 }& \textbf{0.05985} & \textbf{0.15784} & \textbf{0.12288} & \textbf{0.03506} & \textbf{0.05780 }& \textbf{0.05913} &\textbf{0.28605} \\
\hline
LSQ &4  & 0.21309 & 0.10925 & 0.19889 & 0.25601 & 0.04632 & 0.10354 & 0.09807 & 0.63298 \\
SQuAT &4  & \textbf{0.16224} & \textbf{0.08386} &\textbf{ 0.14872} & \textbf{0.07025} & \textbf{0.03275 }& \textbf{0.05179 }& \textbf{0.06944 }& \textbf{0.36209}
\end{tabular}
\caption{The Sharpness comparison between LSQ and SQuAT on GLUE benchmark. Here, $\rho=0.05$. In this case, the local minimum of SQuAT is flatter than LSQ across all the GLUE tasks as well.}
\label{tab:rho005}
\end{table*}

\end{document}